\documentclass[letterpaper]{article} 
\usepackage{aaai25}  
\usepackage{times}  
\usepackage{helvet}  
\usepackage{courier}  
\usepackage[hyphens]{url}  
\usepackage{graphicx} 
\usepackage{natbib}  
\usepackage{caption} 
\usepackage{algorithm}
\usepackage{algorithmic}
\usepackage{booktabs}

\usepackage{graphicx}
\usepackage{subfigure}
\usepackage{amssymb}
\usepackage{mathrsfs}
\usepackage{multirow}
\usepackage{newfloat}
\usepackage{listings}
\usepackage{amsmath}
\usepackage{xcolor}
\usepackage{enumitem} 
\usepackage{newfloat}
\usepackage{listings}
\DeclareCaptionStyle{ruled}{labelfont=normalfont,labelsep=colon,strut=off} 
\floatstyle{ruled}
\newfloat{listing}{tb}{lst}{}
\floatname{listing}{Listing}
\title{AdaCo: Overcoming Visual Foundation Model Noise in 3D Semantic Segmentation via Adaptive Label Correction}
\author{
    Pufan Zou\textsuperscript{\rm 1,2}\equalcontrib, Shijia Zhao\textsuperscript{\rm 1,2}\equalcontrib,
    Weijie Huang\textsuperscript{\rm 1,2},
    Qiming Xia\textsuperscript{\rm 1,2},\\
    Chenglu Wen\textsuperscript{\rm 1,2}\footnote{Corresponding author.},
    Wei Li \textsuperscript{\rm 3}\footnotemark[2],
    Cheng Wang \textsuperscript{\rm 1,2} 
}
\affiliations{
    \textsuperscript{\rm 1}Fujian Key Laboratory of Sensing and Computing for Smart Cities, Xiamen University, China,\\
    \textsuperscript{\rm 2}Key Laboratory of Multimedia Trusted Perception and Efficient Computing,\\
Ministry of Education of China, Xiamen University, China, 
    \textsuperscript{\rm 3}Inceptio\\
    \{zoupufan, zhaoshijia4567, weijie, xiaqiming\}@stu.xmu.edu.cn,\\
    \ clwen@xmu.edu.cn, liweimcc@gmail.com, cwang@xmu.edu.cn
}

\usepackage{bibentry}

\begin{document}

\maketitle

\begin{abstract}
Recently, Visual Foundation Models (VFMs) have shown a remarkable generalization performance in 3D perception tasks.
However, their effectiveness in large-scale outdoor datasets remains constrained by the scarcity of accurate supervision signals, the extensive noise caused by variable outdoor conditions, and the abundance of unknown objects.
In this work, we propose a novel label-free learning method, \textbf{Ada}ptive Label \textbf{Co}rrection (\textbf{AdaCo}), for 3D semantic segmentation.
AdaCo first introduces the Cross-modal Label Generation Module (\textbf{CLGM}), providing cross-modal supervision with the formidable interpretive capabilities of the VFMs.
Subsequently, AdaCo incorporates the  Adaptive Noise Corrector (\textbf{ANC}), updating and adjusting the noisy samples within this supervision iteratively during training.
Moreover, we develop an Adaptive Robust Loss (\textbf{ARL}) function to modulate each sample's sensitivity to noisy supervision, preventing potential underfitting issues associated with robust loss.
Our proposed AdaCo can effectively mitigate the performance limitations of label-free learning networks in 3D semantic segmentation tasks. 
Extensive experiments on two outdoor benchmark datasets highlight the superior performance of our method.
\end{abstract}

%
\section{Introduction}
As a core task of 3D perception, 3D semantic segmentation intends to identify different objects and parts, which has made significant progress in several fields such as robotics, autonomous driving, and mixed reality.
Existing advanced methods \cite{hu2020randla,zhu2021cylindrical} inevitably rely on precise point-wise semantic annotations to train a well-performance segmenter.
However, large-scale annotation for outdoor scenes is time-consuming and labor-intensive. 

In the past few years, many methods adopt weakly supervised learning \cite{hu2022sqn,liu2021one, OPOCA}, semi-supervised learning \cite{Li_Sun_Wu_Song_2021}, and other methods \cite{zhang2023growsp} to further reduce the need for semantic annotation. However, these methods still rely on a certain amount of high-quality 3D annotations.
\begin{figure}[!t]
    \centering  
    
    \includegraphics[width=\linewidth]{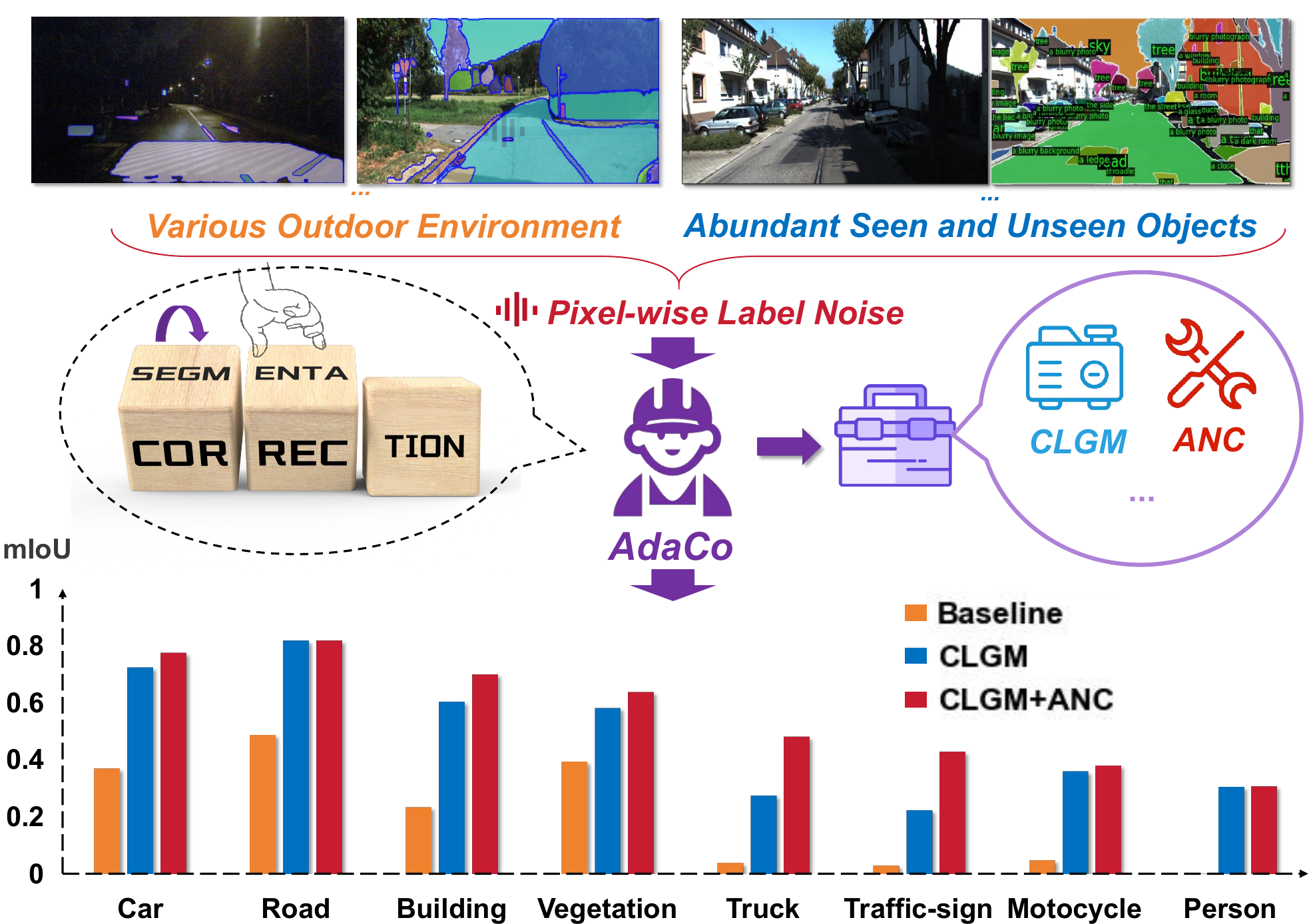}
    

    \caption{
    In outdoor scenes, variable environmental conditions and a rich variety of object categories introduce significant noise when VFMs are applied to 3D perception tasks. 
    Adaco effectively improves the label quality of the baseline method, MaskCLIP, in major categories in road scenes.
    }
  \label{fig:teaser}
\end{figure}


With the powerful understanding and generalization ability of the Visual Foundation Models (VFMs) in zero-shot 2D semantic segmentation, especially SAM \cite{kirillov2023segment} for segment masks and CLIP \cite{radford2021learning} for semantic labels, recent works \cite{chen2023clip2scene,peng2023openscene,chen2023towards} show promising advances in label-free 3D semantic segmentation. Technically, these works generate 3D semantic labels by unprojecting 2D semantic features, extracted by VFMs, onto the corresponding point cloud. 

However, these VFMs-assisted label-free methods also suffer some challenges. 
(1) \emph{Inaccurate pseudo labels.} Current approaches follow the fashion of first extracting semantics from 2D images using VFMs and subsequently embedding them in 2D/3D features. Though refinements based on label consistency and spatial-temporal cues are well explored, the 3D pseudo labels remain inaccurate with great room to improve, particularly for challenging labels such as `truck' and `person' as indicated by the yellow bar in the bottom chart of Fig.~\ref{fig:teaser}.
(2) \emph{Unexpected label noise.} As image-to-point calibration errors cannot be fully eliminated, transferring 2D labels from VFMs into 3D may cause problems like edge blurring. Besides, the influence of illumination, occlusion, and other factors on the image may also introduce noise in the labels.
(3) \emph{Unrobust training process.} If the pseudo labels incorporate significant noise, the 3D segmenter is prone to fitting the noise during training, leading to an effectively unusable model in practical applications.



To address the above challenges, we propose a novel label-free 3D semantic segmentation method, AdaCo, which aims to refurbish noisy labels generated under the guidance of VFMs while alleviating the impact of error samples on the network. 
Firstly, we propose a Cross-modal Label Generation Module (CLGM) to generate 3D point annotations. Specifically, unlike embedding 2D labels into 3D features to yield pseudo labels in previous works, the CLGM lifts VFMs-produced descriptions into 3D space for voting. Without information loss in embedding, the CLGM generates more decent 3D pseudo labels than our baseline, shown in Fig.~\ref{fig:teaser}.
Secondly, to alleviate the noise issue in the pseudo labels caused by the calibration errors and imaging variation, we propose an Adaptive Noise Corrector (ANC) to iteratively correct the mislabeled and fill the unlabeled point-wise semantic annotation to clean samples for the network training.
Lastly, we further combine ANC with an Adaptive Robust Loss (ARL) to reduce the sensitivity of the 3D segmenter to noisy samples during the training process.



Extensive experiments on widely used datasets SemanticKITTI \cite{behley2019iccv} and nuScenes \cite{nuscenes2019} demonstrate the superior performance of our AdaCo over the current baselines.
The main contributions of this work are as follows:
\begin{itemize}
\item We propose a label-free method, Adaptive Label Correction (\textbf{AdaCo}), for 3D semantic segmentation. AdaCo outperforms state-of-the-art label-free 3D segmenters by a large margin.
\item We propose a Cross-modal Label Generation Module (\textbf{CLGM}), leverages VFMs to generate initial noisy 3D pseudo labels from 2D images. Subsequently, we propose the Adaptive Noise Corrector (\textbf{ANC}) to refurbish these noisy labels iteratively.
\item We propose Adaptive Robust Loss (\textbf{ARL}) to alleviate the network's tendency to fit the residual noisy labels, enforcing the penalties for mislabeled labels by the composite loss function.

\end{itemize}

\section{Related Work}
\paragraph{Cross-Modality Supervision for LiDAR Semantic Segmentation.}
3D semantic segmentation tasks typically require high-quality point-wise annotations, which can be labor-intensive and costly. 
To this end, 2D3DNet \cite{genova2021learning} utilizes 2D pseudo labels derived from a pre-trained 2D segmenter to train a 3D segmenter, and apply a credible sparsity filter to deal with noisy 2D pseudo labels. 
2D VFMs such as CLIP \cite{radford2021learning} and SAM \cite{kirillov2023segment}, leveraging their powerful image understanding and generalization capabilities, are also employed to produce higher-quality 2D pseudo labels for label-free 3D semantic segmentation. 
OpenScene \cite{peng2023openscene} and CLIP2Scene \cite{chen2023clip2scene} utilize the CLIP model to generate embeddings between images, text, and points, and design semantic consistency regularization to select positive and negative point samples, training the 3D network using contrastive loss. Chen et al. \cite{chen2023towards} use SAM to refine the CLIP-predicted 2D pseudo labels as 3D cross-modal noisy labels and mitigate the impact of incorrect labels through latent space consistency optimization. However, their performance remains limited by the noise.

\paragraph{Noisy Label Learning for Deep Neural Networks.}
Noisy label learning enables the robust training of deep neural networks in noisy label settings.
To alleviate the impact of noisy labels, some works \cite{sukhbaatar2014training,bekker2016training,lee2019robust} change the architecture to model the noise transition matrix of a noisy dataset and output the estimated-based label transition probability to improve the network's generalization ability. 
Some other works \cite{xia2020robust,wei2021open} use regularization techniques to regulate the network overfitting to noisy labels during training. Still, the label noise rate restricts the model's accuracy. 
\cite{huang2020self} propose a self-adaptive training algorithm to correct problematic training labels by model predictions dynamically. \cite{zheng2020error} propose the LRT method and test the confidence level of the noisy classifier to determine whether a label is correct. \cite{ye2021learning} introduce the label corrector to refurbish the noisy ground truth in the PCSS. However, these methods remain limited for complex open scenes, and the number of network warm-up rounds is manually preset. Our proposed ANC and ARL self-adapt to the learning situation of different samples, continuously refurbish the supervision signals and learn clean samples while mitigating noise label fitting.
\begin{figure*}[htbp]
    \centering
    \includegraphics[width=0.9\linewidth]{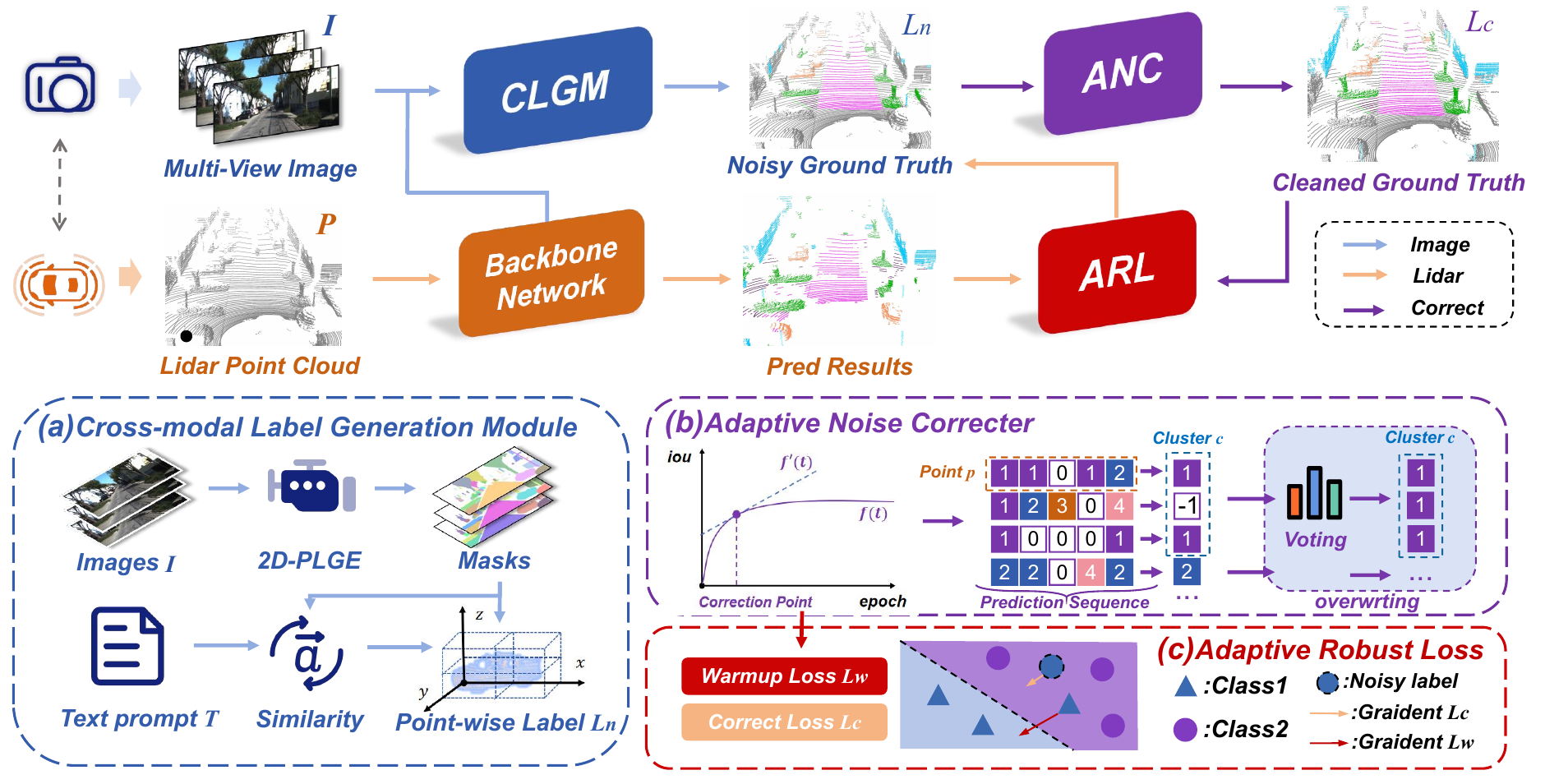}
    \caption{
    The illustration of our label-free 3D semantic segmentation method AdaCo. (a) The 3D noisy pseudo labels are obtained by transferring the VFMs-generate pixel-wise annotation to each point based on CLGM. (b) These noisy pseudo labels are then adaptively refurbished by the proposed ANC. Furthermore, (c) ARL is employed to adaptively adjust the loss method based on the training IoU curve and enforce different penalties for noisy and clean labels.
    }
    \label{fig:pipeline}
\end{figure*}

\section{Methodology}
As shown in Fig. \ref{fig:pipeline}, our AdaCo comprises three modules: (1) CLGM utilizes VFMS for image semantic understanding and labeling 3D point clouds; (2) ANC refurbishes noisy labels using historical predictions and density-based clustering; (3) ARL applies constraints to the network to prevent overfitting to noisy labels. Both ANC and ARL adaptively adjusted according to sample learning.
\subsection{Cross-modal Label Generation Module}
Many works \cite{chen2023clip2scene,peng2023openscene,chen2023towards} use VFMs like CLIP or SAM for enhancing the feature extraction of 2D semantics.  Motivated by their success, we introduce them to point-wise semantic annotations in 3D scenes.
In this case, we propose the Cross-modal Label Generation Module (CLGM) to provide the point-wise pseudo label, serving as the exclusive supervision in network training while identifying more abundant semantic categories.

\begin{figure*}[htbp]
  \centering
  \includegraphics[width=\linewidth]{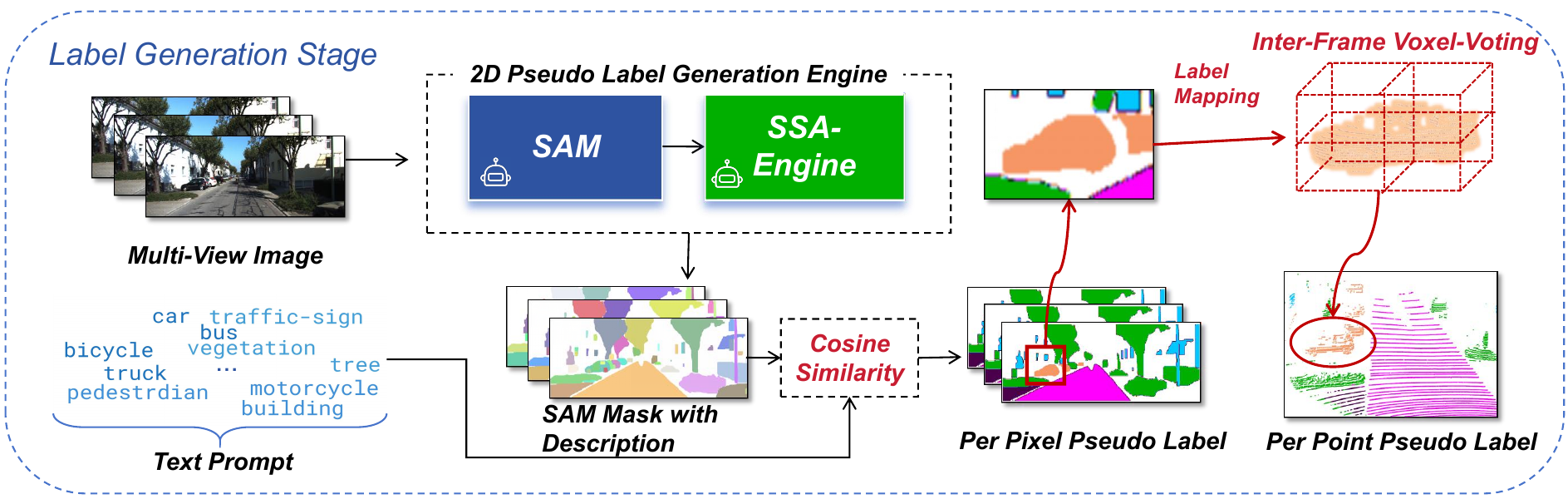}
  
  \caption{CLGM pipeline. We utilize the composition of SAM and SSA-Engine as our 2D-PLGE to segment masks with semantic description, and then calculate the semantic similarity between text prompt and description word by word, 
  and map the class corresponding to the highest semantic similarity to each pixel as its label.
  Finally, we transfer the pixel-wise label to the point after refined by inter-frame voxel-voting. }
  \label{fig:label_generation}
\end{figure*}

\paragraph{Pixel-wise Annotation Generation.}
As shown in Fig.~\ref{fig:label_generation}, we first propose the 2D Pseudo Label Generation Engine (2D-PLGE) to generate pixel-wise pseudo label.
Specifically, take as input multi-view images $\mathcal{I}\in\mathbb{R}^{\mathcal{K}\times 3\times H\times W}$, we obtain a set of class-agnostic binary masks $\mathcal{M}^{\mathcal{K}}=\left\{m^{\mathcal{K}}_1, m_2^{\mathcal{K}}, ..., m^{\mathcal{K}}_n|m^\mathcal{K}_n \in \{0,1\}^{W \times H}\right\}$ with the assistance of the FastSAM \cite{zhao2023fast}, 
where $m_n^{\mathcal{K}}$ denoting the $n$-th mask in the $K$-th view image, $W$ and $H$ are the width and height of the image, respectively.
For each mask $m^{\mathcal{K}}$, we fed it along with its corresponding image patch into the SSA-Engine \cite{chen2023semantic} to generate the semantic description $\mathcal{D}^\mathcal{K}=\left\{D^{\mathcal{K}}_1, D_2^{\mathcal{K}}, ..., D^{\mathcal{K}}_n\right\}$.
Then, we utilize the category vocabulary as a textual prompt $T$ to calculate the semantic similarity between $T$ and the generated descriptions.
Ultimately, we select the semantic class with the highest similarity to the text prompt as the final annotation $y^{k}_{2D}$ for each image pixel.

\paragraph{Point-wise Label Unprojection and Voxel Refinement.}
By utilizing the existing pixel-to-point calibration relationships and matrix computations, we derive point-wise annotations $y_{3D}$ by back-projecting pixel-wise annotations $y^{k}_{2D}$ from different viewpoints onto the point cloud.
Leveraging the continuity of camera capture and point cloud acquisition across frames, we voxelize the point cloud and align the adjacent frame's point cloud to the current frame's coordinate system using the calibration matrix. We then vote the semantic annotation of the back projection point by point, to initially refine the quality of the pseudo label.

\subsection{Adaptive Noise Corrector}
Issues like lens distortion and sampling rate which lead to blurred object edges and the occlusion problem also introduce extensive noise. To address this, we designed an Adaptive Noise Corrector (ANC) that utilizes the network’s early learning with clean samples and point cloud clustering constraints.

\begin{figure}[t]
  \centering
  \subfigure[Training mIoU]{
  \includegraphics[width=0.46\linewidth]{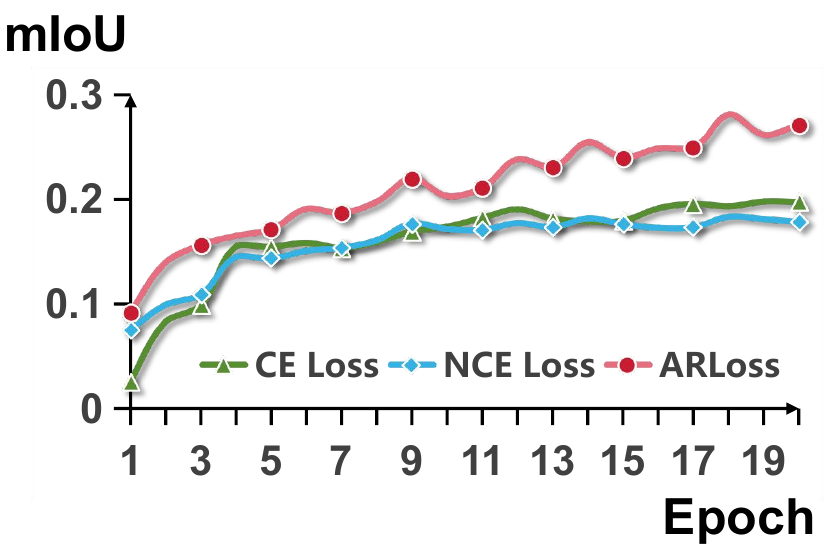}
  }
    \subfigure[Early Learning mIoU]{
  \includegraphics[width=0.46\linewidth]{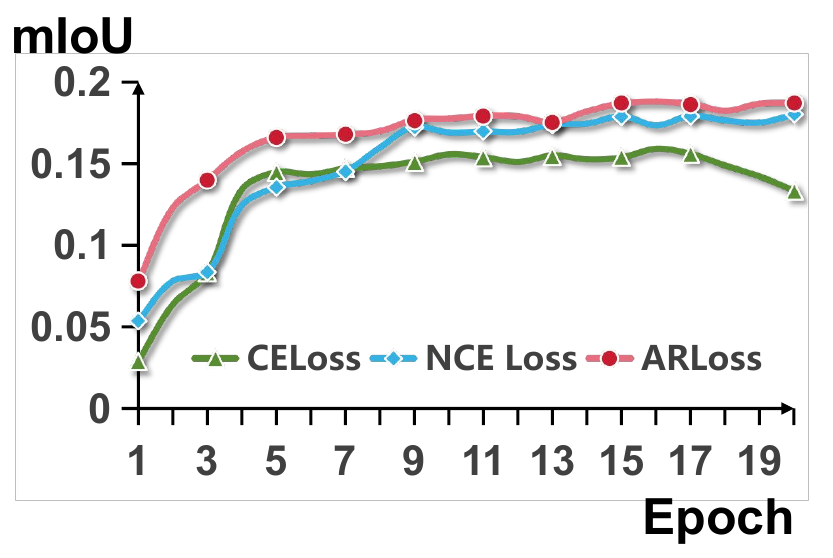}
  }
   
   \caption{
   The learning curve of different mIoU in the SemanticKITTI \textit{train} set.
   The training mIoU is calculated with the noisy ground truth, while the early learning IoU curve is calculated with the correct ground truth.   }
   \label{fig:learning_curve}
\end{figure}

\paragraph{Adaptive Correction Timing Search.}
Inspired by \cite{liu2022adaptive}, we first harness the early learning capabilities of deep neural networks to extract knowledge from clean data samples. This process will continuously update the noisy labels by the network's historical predictions on unknown samples and dynamically adjust warm-up iterations according to each sample's learning curve.
As the green curve in Fig.\ref{fig:learning_curve} shows, monitoring training mIoU and early learning mIou on the training set, we note a rapid early increase in both, followed by a slowdown or decline at the same time as the model memorizes noise. 
Therefore, this turning point can be regarded as the pivotal moment at which each sample achieves saturation in learning from the clean data.
For each iteration $t$, we let each sample $s$ from the sample set $S$ record its training mIoU, and then use the least squares method to fit an exponential function $f$ to the mIoU as:
\begin{equation}
\label{iou_fit_func}
    \textit{f}_s(t) = a_s \cdot \left(1 - \exp\left(-\frac{1}{c_s} \cdot t^{b_s}\right)\right), \quad \forall t \in T, \forall s \in S
\end{equation}
where $a$, $b$, and $c$ are the fitting parameters. Subsequently, we calculate the derivative of the fitting function $f'$ with respect to $t$ from 1 to the current iteration $t_c$. We then set a threshold $r$. 
When the change in the derivative exceeds this threshold $r$, we consider it the moment $t_c$ to correct the annotation.
\paragraph{Reliable Labels Prediction.}
After a correction round, $t_c$, we analyze the network’s historical predictions for each instance,
using the mode of these predictions to refurbish the noise labels.
Additionally, following PNAL \cite{ye2021learning}, we consider samples with consistent historical predictions and their labels as reliable.
The point $x_s$ in sample $s$ is defined as a reliable sample if its confidence in historical prediction satisfies $F(x_s;q)\geq\gamma(0\leq\gamma\leq1)$. The predictive confidence is defined as:
\begin{equation}
    F(x_s;q)=-\frac{\text{entropy}(P(k|x_s;q))}{\log{\frac{1}{K}}},
\end{equation}
where $q$ is the length of history predictions. 
We hope that the length of the historical prediction record $q$ can be self-adapted according to the sample, that is, the number of training rounds before the correction round $t_c$ occurs, but $q$ will not exceed the upper bound $t_m$ to save memory used during training. The label history $H_{x_{s}}(q)$ of the point $x_s$ denotes the predicted labels of the previous $q$ rounds. The predicted labels of the points at time $t$ as $\hat{y}_{t}$, and $H_{x_{s}}(q)=\left\{\hat{y}_{t1},..., \hat{y}_{tq}\right\} $. 
For the point $x_s$ in sample $s$, we denote its output probability of the label $k$ as $P(k|x_s;q))$, where $k\in \left\{1,...,K\right\}$.
We can calculate the probability as:
\begin{equation}
    P(k|x_s;q)) = \frac{\sum_{\hat{y}\in H_{x_{s}}(q)}[\hat{y}=k]}{q},
\end{equation}
\begin{equation}
    q=t_c \quad \text{if } t_c < t_m, \quad t_m \quad \text{otherwise},
\end{equation}
where $[\cdot]$ is the Iverson bracket. We define the set of reliable points in sample $s$ as $X_\mathrm{reliable}^{s}$.
For each point in the reliable set, we calculated its reliable labels as:
\begin{equation}
 y_{x_s}^{*} = \text{argmax}P(k|x_s;q).
\end{equation}
\paragraph{Clustering Label Propagation.}
We employ the DBSCAN \cite{ester1996density} to provide instance information within the sample, addressing the issue of edge blur caused by direct back-projection. Initially, we remove ground points from the point cloud scene and apply the DBSCAN clustering algorithm to the remaining non-ground points to form $m$ clusters $C=\left\{c_1,...,c_m\right\}$. 
For each cluster, we replace the labels of the intersection points with the set of reliable points $X_{\text{reliable}}^{s}$. The new labels $L_{ref}$ are determined by a voting mechanism within the intersection area.
We have counted the frequency of each label within the intersection area and constructed a label frequency vector $\text{Fre}_s = \left\{\text{Fre}_s^{0},..., \text{Fre}_s^{K}\right\}$ based on this information, where $\text{Fre}_s^{k}(0\leq k \leq K)$ represents the frequency of class $k$ in this intersection area. The final winner label $L_{ref}$ is randomly selected under the guidance of this vector, where $L_{ref}=\text{Random}\left\{k|\text{Fre}_s^{k} \geq \frac{max(\text{Fre}_s)}{\omega}\right\}$. 
Unlike the method used in the \cite{ye2021learning}, after updating a portion of the labels, 
we enhance the network's memory for clean samples by applying the complete label information from the entire sample set during gradient calculation. The Algorithm \ref{algo1} shows the process of the Adaptive Noise Corrector.

\subsection{Adaptive Robust Loss}

The robust loss function is crucial for training with noisy labels. However, according to \cite{ma2020normalized,zhang2018generalized,wang2019symmetric}, it may hinder optimal convergence and early learning,  potentially causing underfitting. On the other hand, cross-entropy loss offers rapid noise-free label early learning ability. Therefore, we propose optimizing training with two adjustable loss functions, guided by sample learning. In the initial phase of training, before the correction round $t_c$, our loss function primarily comprises cross-entropy loss on the 3D network and mean squared error loss between feature $Z_{2D}$ from 2D knowledge branch and $Z_{C}$ from camera branch following \cite{cen2023cmdfusion}. The ARL in the warmup stage is defined below:
\begin{equation}
\begin{split}
    L_{\text{warmup}} &=L_{\text{CE}}+L_{\text{MSE}}( Z_{2D},Z_{C})+L_{\text{Lov\H{a}sz}}(y,\hat{y}),\\
\end{split}
\end{equation}
where $\hat{y}$ is the prediction of 3D
network, $L_{\text{Lov\H{a}sz}}$ denotes the Lovasz loss function \cite{berman2018lovasz}.

Once the correction operation is initiated, we introduce additional terms to the loss function, including normalized cross-entropy loss and mean absolute error loss following \cite{ma2020normalized}, while diminishing the dominant weight of the preceding loss functions.  The normalized cross-entropy loss and mean absolute error loss are as follows:
\begin{equation}
\begin{split}
    L_{\text{NCE}} &=\frac{\sum_{k=1}^{K}q(k|x)\log{p(k|x)}}{\sum_{j=1}^{K}\sum_{k=1}^{K}q(y=j|x)\log{p(k|x)}},
\end{split}
\end{equation}
where the $x$ denotes a sample in the dataset, $K$ is the number of class. $q(y=k|x)$ denotes the distribution  over different labels in class $k$ for sample $x$, and the $\sum_{k}^{K}q(y=k|x)=1$. The $p(k|x)$ denotes the probability output of the network as $p(k|x) = \frac{e^{z_{k}}}{\sum_{j=1}^{K}e^{z_{j}}}$,$z_{k}$ represents the logits output of the network in class $k$.
\begin{equation}
    L_{\text{MAE}} = \frac{1}{N} \sum_{i=1}^{N} \left| y_i - \hat{y}_i \right|,
\end{equation}
where $N$ presents the number of points. The additional loss term is calculated as:
\begin{equation}
    L_{\text{corr}}= \lambda L_\mathrm{NCE}+\beta L_\mathrm{MAE},
\end{equation}
where the $\lambda$ we set it to 100, and $\beta$ we set it to 1 as \cite{ma2020normalized} suggests.
The overall adaptive robust loss $L_{ARL}$ for our training model is calculated as:
\begin{equation}
\begin{split}
    L_{\text{ARL}} = \begin{cases}
        L_{\text{warmup}}&t<t_c \\
       L_{\text{corr}}+L_{\text{warmup}}+\sigma L_{\text{CE}}&t\geq t_c,
    \end{cases} 
\end{split}
\end{equation}
where the $\sigma$ we set to be -0.99 to reduce the overfitting of the cross-entropy function to noise while ensuring its sensitivity to gradient updates.

\section{Experiements}

Aiming to evaluate the robust performance of our proposed AdaCo in the environment of noisy annotations, we conduct experiments on two large-scale outdoor LiDAR scene datasets, SemanticKITTI \cite{behley2019iccv} and nuScenes \cite {nuscenes2019}, and compare our method with the previous 3D label-free semantic segmentation methods.
At the same time, we conducted a complete ablation experiment to verify the feasibility of each module. We will elaborate on the complete process of our experiments in terms of dataset, training details, ablation study, etc.

In our experiments, we utilize the CMDFusion \cite{cen2023cmdfusion} backbone network as our baseline, which is guided by pixel-wise pseudo labels generated from multi-view images in each dataset using MaskCLIP \cite{zhou2022extract}.

\begin{algorithm}[H]
    \setlength{\algorithmicindent}{0.5em}
    \caption{Adaptive Noise Corrector} 
    \label{algo1} 

    \begin{algorithmic}[1]
        \REQUIRE LiDAR points  $X_{s}$ of sample $s$, the noisy label$~L_{s}$, $K$ classes, The clusters$~C$;
        \ENSURE The refurbished label $L_{ref}$;
        
        \STATE $H_x(q) = [\;],I_s = [\;]$
        \FOR {$\mathrm{epoch}~ t = 0, 1, 2, ..., T$}
            \FOR{$\mathrm{sample} ~s \in S$}
            \STATE $\hat{y}_s = \mathrm{Model}(X_s)$
            \STATE $I_s$.append$\left( \mathrm{mIoU}\left(\hat{y}_s,L_{s}\right)\right)$\STATE $f(t) = \mathrm{fitFunc}\left(I_s\right)$
            \IF{$\frac{|f'(1)-f'(t)|}{f'(1)} > r$}
                \STATE $X_{\text{reliable}}^{s} = \left\{x_{s}|F(x_s;q)>\text{threshold}\right\}$
                \STATE $y_{\text{reliable}}^{s}=\text{argmax}(\frac{\sum_{\hat{y}\in H_x(q)}[\hat{y}=k]}{|H_x(q)|})$
                \STATE $C_{i^{*}} = \left\{x_s|~\exists x_s \in X_{\text{reliable}}^{s} \cap x_s \in X_{C_i},C_{i}~\in C\right\}$  
                \STATE$\text{Fre}_{C_{i^{*}}}^{k}=\sum{[y_{\text{reliable}}^{s}=k][x_s \in X_{C_i}]}, \;(k\in[1, 2, \ldots, K])$
                \STATE$L_{ref}=\mathrm{Random}\left\{k|\text{Fre}_{s}^{k}\geq \frac{max(\text{Fre}_s)}{\omega}\right\}$
                \STATE...
            \ENDIF
            \STATE$H_x(q).append\left( \hat{y}_s\right)$
            \ENDFOR
        \ENDFOR 

    \end{algorithmic} 
\end{algorithm}


\paragraph{Datasets and Metrics.}

SemanticKITTI \cite{behley2019iccv} has 22 sequences with 20 semantic classes. We use 19,130 samples for training from sequences 00-10, and 4,071 samples for validation from sequences 08. The dataset provides 2 images of the front-view camera for each sample. 

The nuScenes \cite{nuscenes2019} dataset is widely used in 3D perception tasks \cite{coin, hinted}. It contains 28,130 samples for training, 6,019 samples for validation, 6,008 samples for testing, and 17 semantic classes. Each sample has LiDAR data of 32 beams and RGB images captured by 6 cameras, covering the whole 360 angle.

We present the results with the official evaluation metrics: mean Intersection over Union (mIoU) and mean Accuracy (mAcc). All experiments are conducted on point clouds and not on images in both datasets.

\setlength{\tabcolsep}{1mm}
\begin{table*}[htbp]

    \label{tab:mainfigure}

\centering
\begin{tabular}{c|c|c|c|c}
\toprule
\multirow{2}{*}[0pt]{Settings}&\multirow{2}{*}[0pt]{Methods}& \multirow{2}{*}[0pt]{Publication} & \multicolumn{2}{c}{mIoU(\%)}  \\
&&&nuScenes&SemanticKITTI\\
\midrule
\multirow{6}{*}[0pt]{Label-Free}&MaskCLIP~\cite{zhou2022extract}        & ECCV 2022 & 12.8&- \\
&MaskCLIP++~\cite{zhou2022extract}      & ECCV 2022 & 15.3&- \\
&CLIP2Scene~\cite{chen2023clip2scene}   & CVPR 2023 &20.8&- \\

&OpenScene~\cite{peng2023openscene}   &  CVPR 2023& 14.6&- \\

&TowardsVFMs~\cite{chen2023towards}     & NIPS 2023 & 26.8&-\\

&Hierarchical~\cite{kang2024hierarchical}   &  CVPR 2024 & 23.0&- \\
\midrule
\multirow{2}{*}[0pt]{Unsupervised}&GrowSp~\cite{zhang2023growsp}     & CVPR 2023 &-&13.2\\

&U3DS\textsuperscript{3}~\cite{liu2024u3ds3}     & CVPR 2024&- &14.2\\
\midrule
\multirow{2}{*}[0pt]{Label-Free}&Baseline (MaskCLIP$^\dagger$)~\cite{zhou2022extract}& - & 16.5&8.1 \\

&Ours &-&\textbf{31.2} &\textbf{25.7}  \\
        
\bottomrule
\end{tabular}
\caption{Comparison results on official SemanticKITTI single scan validation set and the nuScenes val set with current state-of-the-art label-free and unsupervised methods for point cloud semantic segmentation tasks.  $\dagger$ represents the reproduced result. }
\label{tab:kitti}
\end{table*}

\paragraph{Implementation Details.}
Our backbone network settings follow the CMDFusion \cite{cen2023cmdfusion}.
We set voxel size to [0.05m,0.05m,0.05m] for SemanticKITTI and [0.1m,0.1m,0.1m] for nuScenes. The 2D-PLGE is kept frozen for inference. To obtain shapes of instances in the image that are as complete and non-overlapping as possible, we set the segmentation threshold of FastSAM \cite{zhao2023fast} to 0.7 and the mask confidence value to 0.4. For the voxel size of inter-frame voting, we adopted the same settings as those in \cite{cen2023cmdfusion}.
In the CLGM, we utilize dataset categories as text prompts and apply word2vec \cite{church2017word2vec} to assess the word-by-word similarity between the 2D-PLGE-generated mask descriptions and the prompts.
In the ANC, we set the derivative threshold $r$ for label correction to 0.9, and our refinement is conducted at the sample level. 
Each sample records its mIoU for each round of training to fit a curve, whose parameters satisfy 0 $ \leq a \leq $1, b $\geq$ 0, c $\geq$ 0. We record the network’s prediction results for each round of the sample to form a historical prediction sequence, with the maximum sequence length set to 5 due to memory constraints and they only correct once during the training phase to avoid the label quality decrease. 
The parameter $\omega$ we choose the winner label in the cluster is set to be 3. In the ARL, we set the parameters $\lambda$ and $\beta$ to be 100 and 1 as \cite{ma2020normalized} suggests, emphasizing more active learning and less passive learning. The weight $\sigma$ for minimizing cross-entropy loss is set to -0.99, and we have conducted various experiments to determine the optimal parameter settings. We train the models using the SGD optimizer with a learning rate of 0.24 for 40 epochs and adopt the cosine annealing warm restart strategy for learning rate decay.  All experiments are trained on 4 RTX 3090 GPUs.

\subsection{Comparison Results}
\begin{figure*}[htbp]
  \centering
  \subfigure[Ground Truth]{
    \includegraphics[width=.23\linewidth]{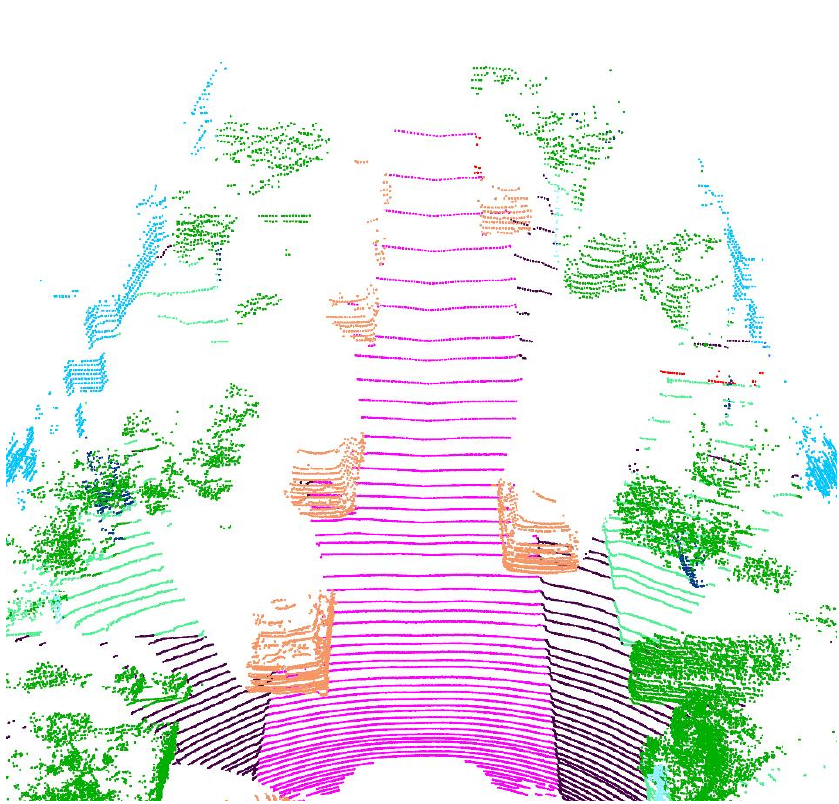}
  }
  \subfigure[Baseline]{
    \includegraphics[width=.23\linewidth]{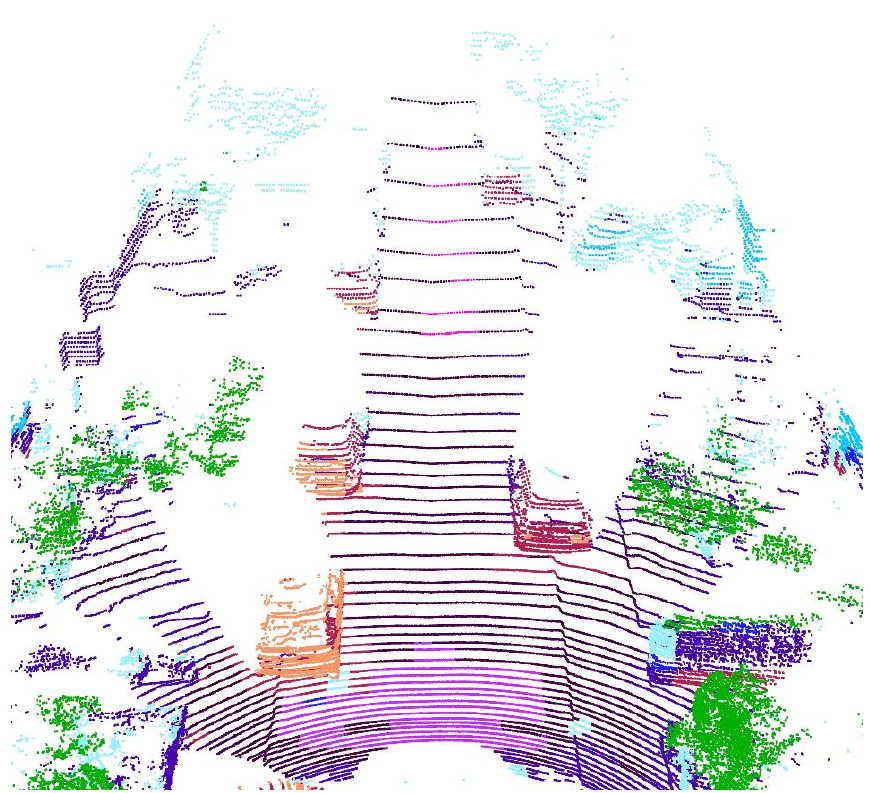}
  }
  \subfigure[Noisy Ground Truth]{
    \includegraphics[width=.23\linewidth]{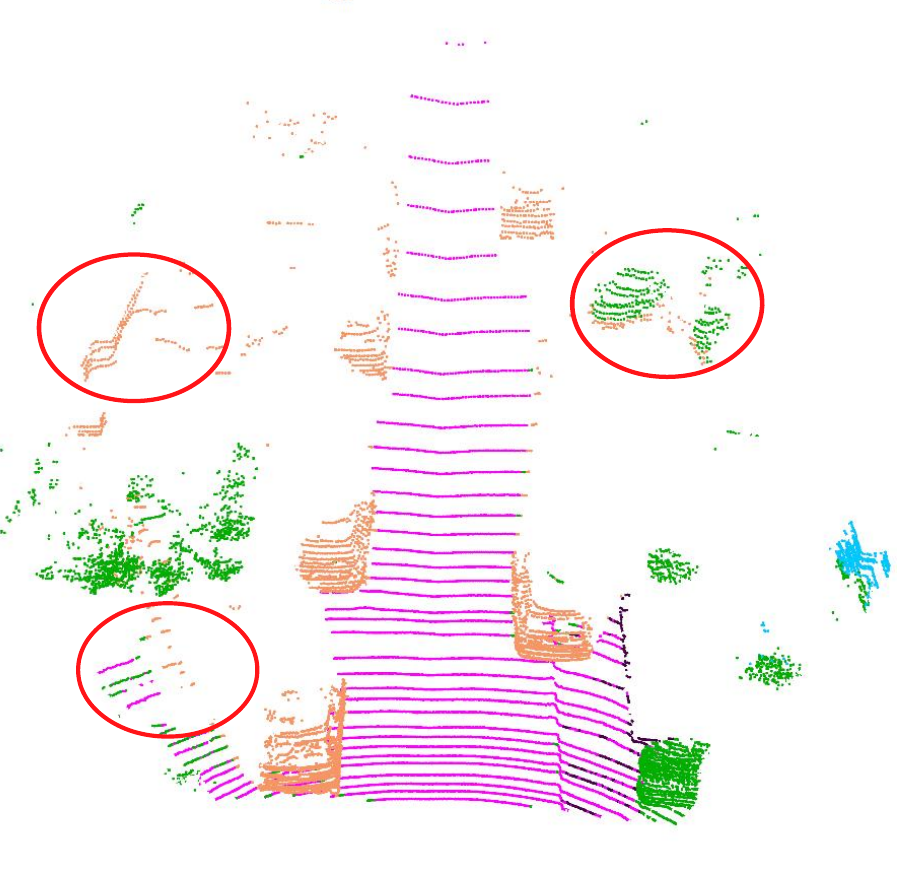}
  }
  \subfigure[Ours]{
    \includegraphics[width=.23\linewidth]{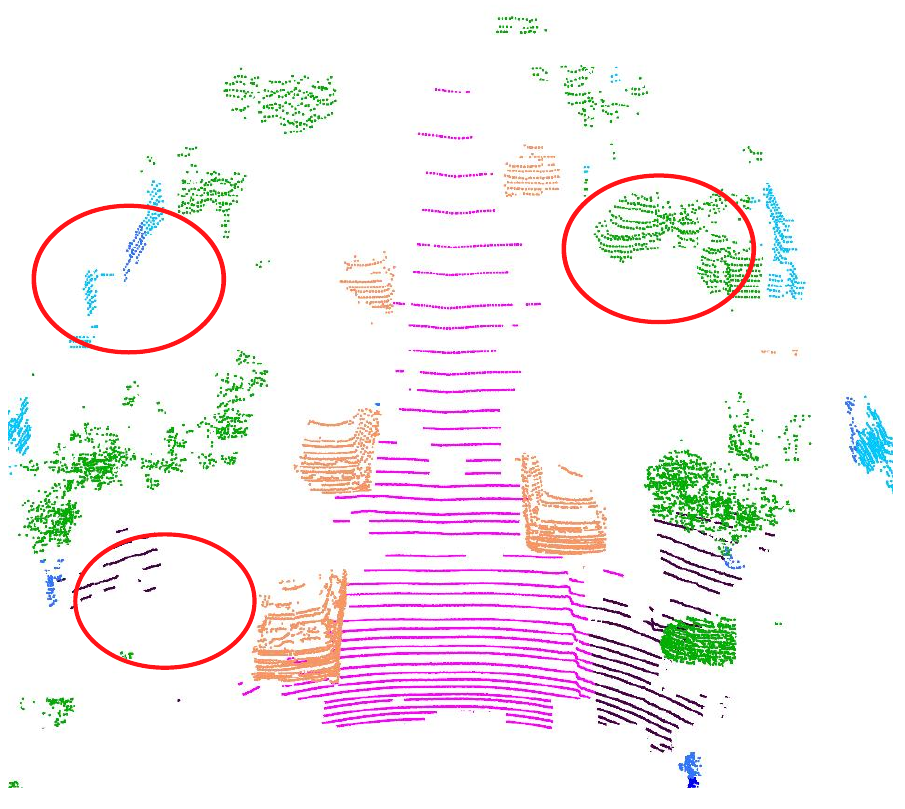}
     }
     
    \subfigure[Ground Truth]{
\includegraphics[width=0.23\linewidth]{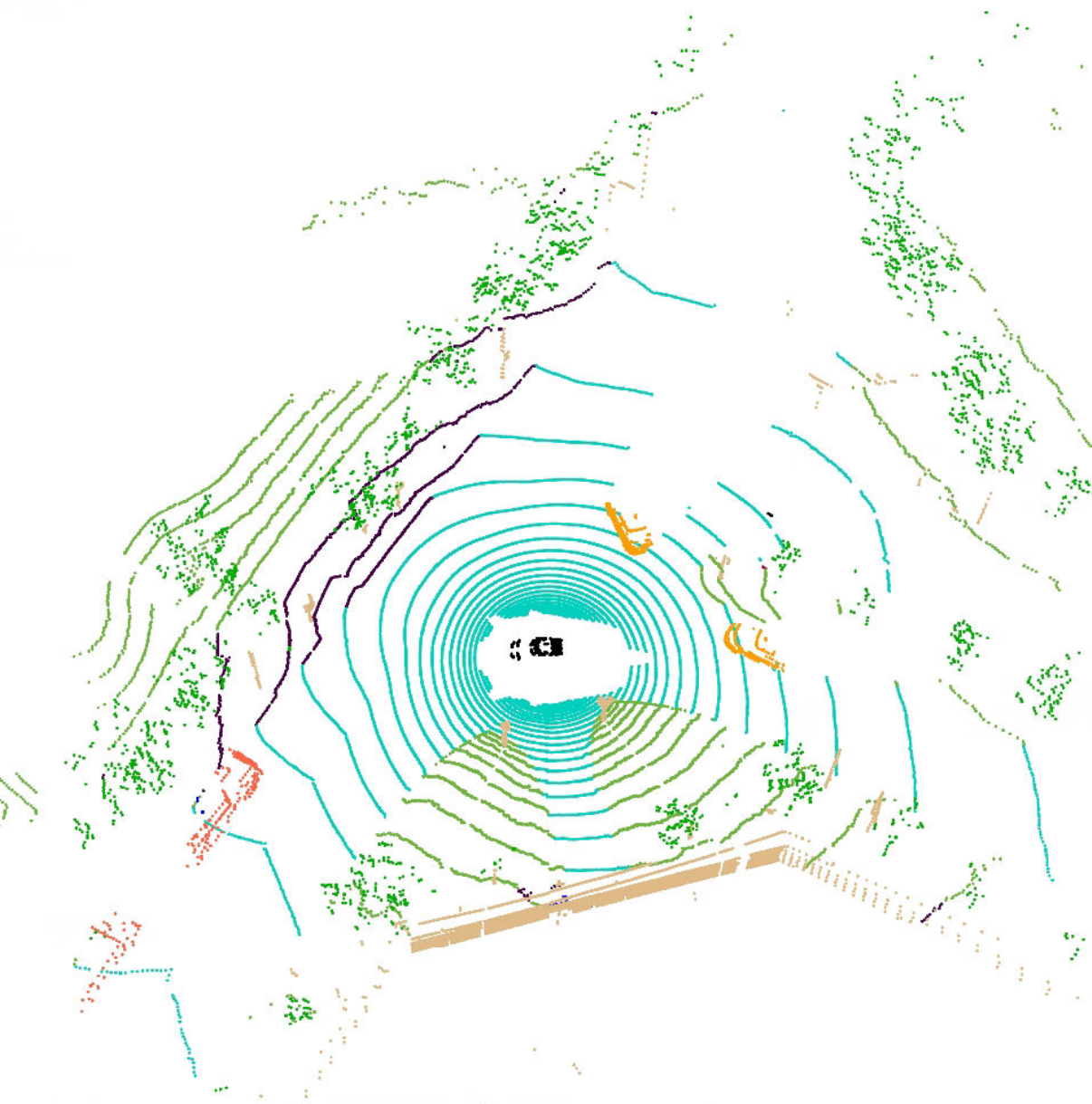}}
\subfigure[Baseline]{
		\includegraphics[width=0.23\linewidth]{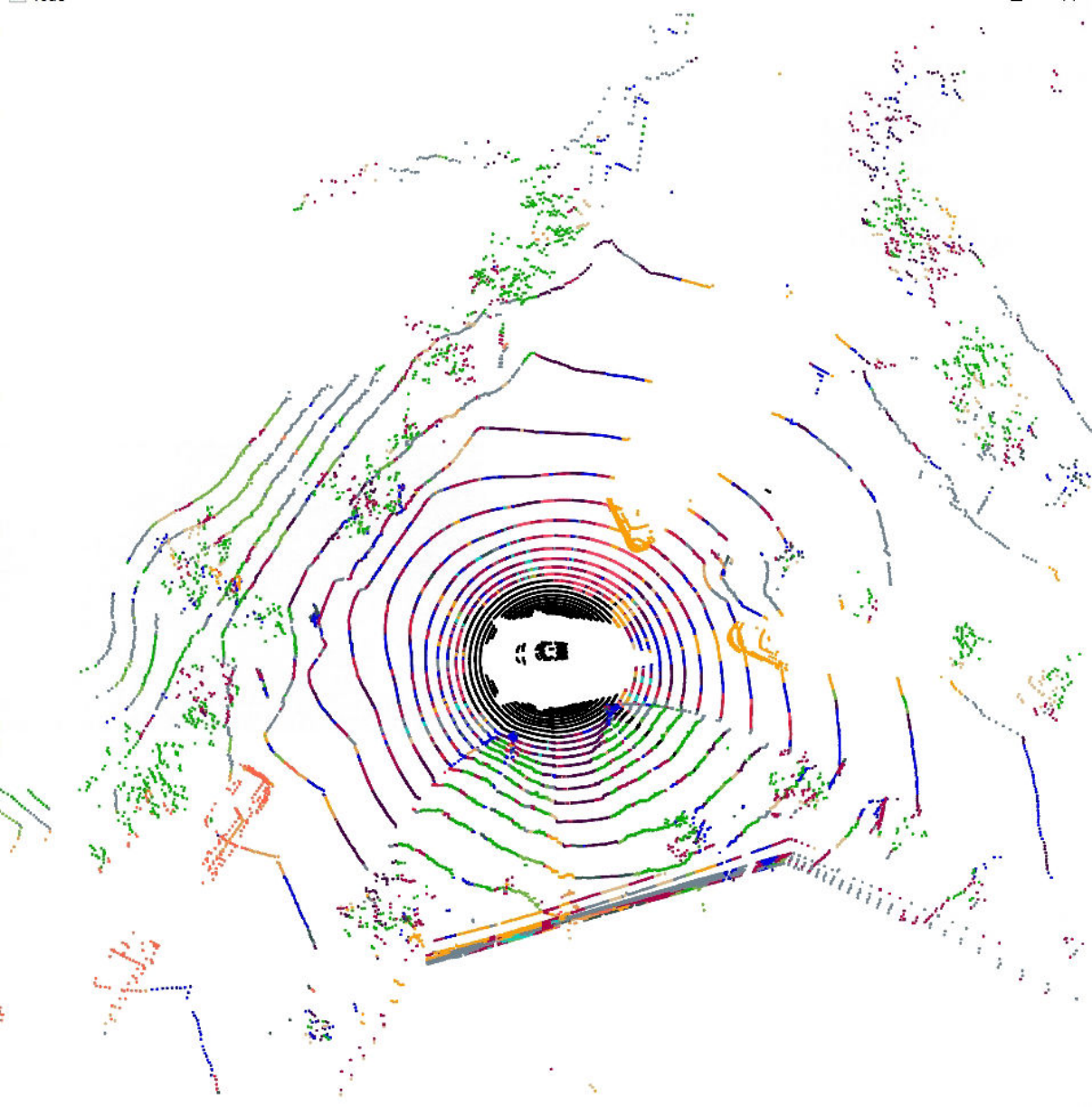}}
\subfigure[Noisy Ground Truth]{
		\includegraphics[width=0.23\linewidth]{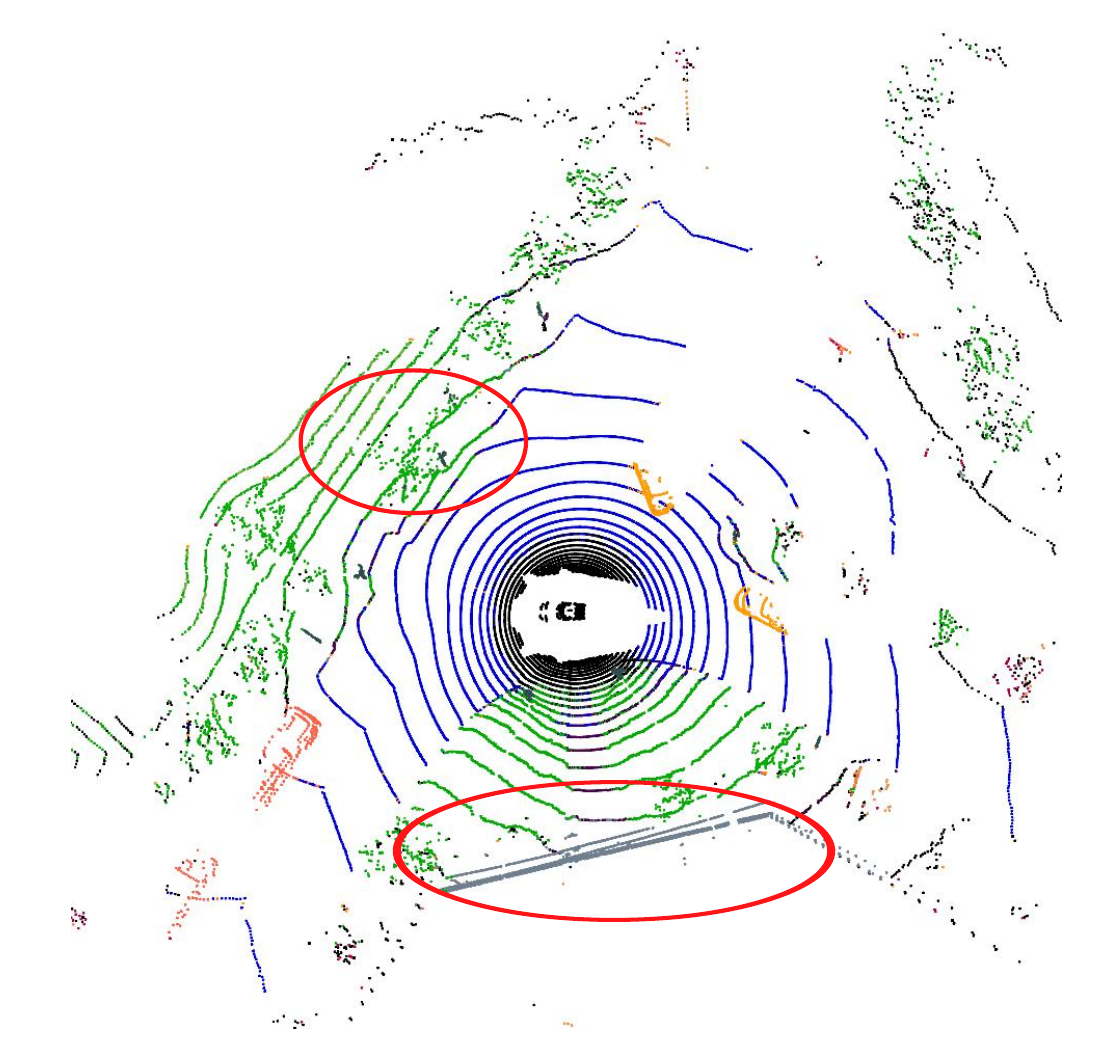}
  }
  \subfigure[Ours]{
		\includegraphics[width=0.23\linewidth]{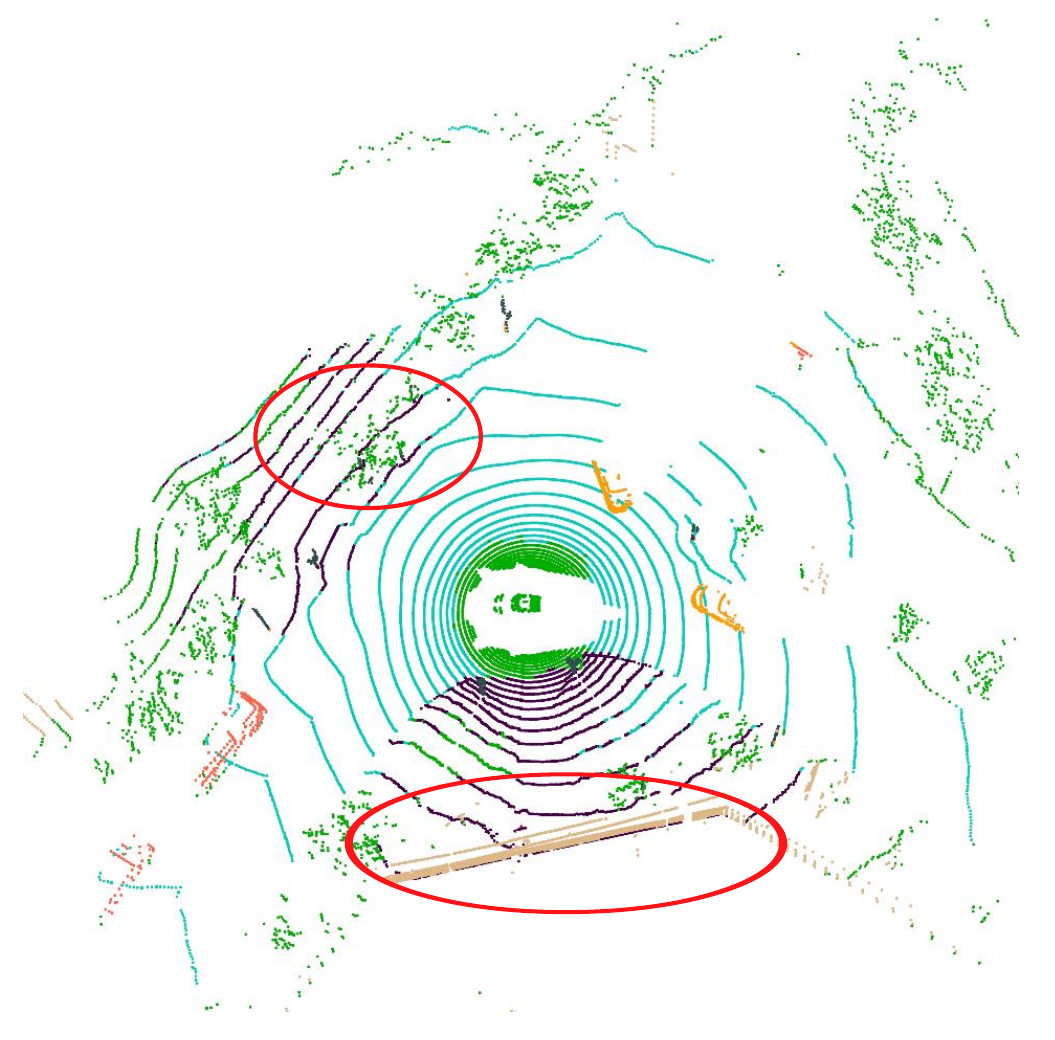}}

  \caption{Qualitative results in SemanticKITTI (a)-(d) and nuScenes (e)-(h), the noisy ground truth is generated from CLGM. As shown by the red circle in the figure, our Adaco remembered clean samples and mitigated the blurring of object edges. }
  \label{fig:qualitative}
\end{figure*}

\paragraph{Results on SemanticKITTI.}
SemanticKITTI has two images of the front-view camera for each sample and we fed them into MaskCLIP to generate the 2D pseudo labels, then we used the pixel-to-point calibration matrix to transfer the pixel-wise pseudo labels to the point and training on CMDFusion \cite{cen2023cmdfusion} as our baseline.   
As shown in Tab.~\ref{tab:kitti}, our AdaCo achieves a 17.6 mIoU improvement over the baseline. 
Furthermore, compared to the unsupervised methods, 
our method respectively achieves 12.5 and 11.5 higher mIoU. Fig.  \ref{fig:qualitative} presents a qualitative comparison of the results between the ground truth, noisy ground truth, baseline, and our AdaCo method.

\begin{table}[ht]
\centering

        \begin{tabular}{c|c|c|c|c}
        \toprule
          CLGM  & ANC & ARL & mIoU(\%)    &mAcc(\%)   \\
        \midrule
        &&&8.1&35.1\\
        
        $\checkmark$&&&19.5&58.2\\
        $\checkmark$&$\checkmark$&&25.3&73.2\\
        $\checkmark$&&$\checkmark$&21.4&65.9\\
        $\checkmark$&$\checkmark$&$\checkmark$&25.7&74.9\\
        
        \bottomrule
       \end{tabular}
       \caption{Ablation study on SemanticKITTI \textit{validation} set. }
        \label{tab:ablation}
\end{table}

\paragraph{Results on NuScenes.}
    In this experiment, we compared the AdaCo method with existing 3D
    label-free semantic segmentation methods, which also rely on 2D VFMs to extract semantic features from images for guiding 3D networks. As shown in Tab.~\ref{tab:kitti}, our method outperforms the state-of-the-art method by 4.4 in mIoU, and the qualitative results in Fig. \ref{fig:qualitative} demonstrate the effectiveness of our approach.
    

        


\subsection{Ablation Study}
To validate the effectiveness of each component in AdaCo, we conducted ablation experiments on the SemanticKITTI validation set and reported the results in Tab.~\ref{tab:ablation}.
In addition, we have also verified the impact of hyperparameters for each component in AdaCo and reported the results in Tab.~\ref{tab:hyperparameters}.


        

\paragraph{Effect of CLGM.} 
As the Tab.~\ref{tab:ablation} indicates, our CLGM demonstrates better semantic comprehension capabilities than the baseline.
In Tab.~\ref{tab:hyperparameters}, the results presented in the \emph{Adjacent Frame} row illustrate the impact of the length of adjacent frames. As the length increased, the quality of the labels improved. Due to memory constraints, we did not use longer adjacent frames to optimize the labels.

\begin{table}[ht]
\setlength{\tabcolsep}{0.4mm}
\small
\centering

        \begin{tabular}{c|c|c|c}
        \toprule
         Ablation Target&Settings& mIoU (\%)   
  &mAcc(\%)   \\
        
        \midrule
        \multirow{2}{*}[0pt]{Adjacent Frame}&1,w/o train&25.4&74.6\\
        &2,w/o train&25.9&75.0\\
        \midrule
        \midrule
        \multirow{3}{*}[0pt]{History Seq.}
        &3&18.5&56.2\\
        &4&21.6&65.5\\
        &5&25.3&73.2\\
        \midrule
        \multirow{2}{*}[0pt]{Adaptive Correction}&Correct Each Down&18.9&57.3\\
        &Correct Once&24.0&70.4\\
        \midrule
        \midrule
        \multirow{5}{*}[0pt]{Robust Loss Weight}
        &$\lambda/\beta=10,\sigma=0$&18.0&54.7\\
        &$\lambda/\beta=10,\sigma=-0.9$&18.1&55.3\\
        &$\lambda/\beta=100,\sigma=-0.9$&18.4&56.1\\
         &$\lambda/\beta=10,\sigma=-0.99$&21.4&65.9\\
        &$\lambda/\beta=100,\sigma=-0.99$&21.6&66.3\\
        \midrule
        &Full&25.7&74.9\\
        
        \bottomrule
       \end{tabular}
       \caption{The impact of the module hyperparameters on SemanticKITTI \textit{validation} set.}
        \label{tab:hyperparameters}
\end{table}

\paragraph{Effect of ANC.}
As shown in Tab.~\ref{tab:ablation}, our ANC strategy significantly enhances the quality of CLGM-generate pseudo labels.
In Tab.~\ref{tab:hyperparameters}, we adjusted the maximum historical prediction record values for various lengths in the \emph{History Seq.} row, suggesting a shorter length would result in a lack of reference for ANC to correct samples. Due to memory constraints and performance considerations, the optimal length was determined to be 5. 
Additionally, the results in \emph{Adaptive Correction} row indicate that adaptive correction only corrected once per sample performed better than correct at each time the derivative of the fitting function started to decrease.

\paragraph{Effect of ARL.}
As depicted in Tab.~\ref{tab:ablation}, ARL effectively mitigates the interference of noisy pseudo labels on the model, demonstrating a more robust segmentation performance.
Moreover, we refine the ARL by adjusting the weights of the various active, passive, and unnormalized cross-entropy losses during the correction phase. 
As shown in the \emph{Robust Loss Weight} row in Tab.~\ref{tab:hyperparameters}, the model performance tended to improve when decreasing the contribution of the unnormalized loss to the total loss.
Besides, the model's resilience to noise is also strengthened after incorporating the additional robust noise loss terms.
Fig. \ref{fig:learning_curve}             shows that ARL has better early learning capabilities in the initial stage of the model and is less prone to fitting noisy samples.

\section{Conclusion}
We propose a novel label-free point cloud semantic segmentation approach, AdaCo, which leverages VFMs to provide cross-modal 3D semantic pseudo labels and adaptively adjusts the noisy labels within the samples.
Furthermore, it robustly learns from noisy information.
Firstly, we employ the CLGM to extract semantics from images and transfer them to point clouds to provide point-wise pseudo labels. Next, we propose ANC and ARL to mitigate the impact of noisy labels provided by the CLGM on network learning. ANC refurbishes the supervision signals by integrating the instance shapes within the scene with historical predictions. ARL employs a combination of normalized cross-entropy loss and robust loss to enhance the penalty for incorrect annotations on network learning.
\textbf{Limitations:} AdaCo relies on pixel-to-point calibration for semantic label transfer and is limited by the semantic understanding of VFMs in driving contexts, potentially failing with small image targets and diverse categories.
Future work may explore decoupling 2D image calibration from 3D point clouds for cross-dataset supervision or offering a more precise semantic capture of road images.
\section{Acknowledgment}
This work was supported by the National Natural Science Foundation of China (No.62171393), and the Fundamental Research Funds for the Central Universities (No.20720220064).
\section{Additional Implementation Details.}
Our backbone network follows the settings of CMDFusion \cite{cen2023cmdfusion}, which is a 3D semantic segmentation network based on SPVCNN \cite{spconv2022}. For different datasets, we set the voxel size to [0.05m, 0.05m, 0.05m] for SemanticKITTI, and [0.1m, 0.1m, 0.1m] for NuScenes. The valid space of the LiDAR point cloud is set to [-50m, -50m, -4m] to [50m, 50m, 2m] for both the training and validation sets in the SemanticKITTI dataset and [-100m, -100m, -4m] to [100m, 100m, 17m] for the nuScenes dataset.
\begin{table}[ht]
\centering
        \begin{tabular}{c|c|c}
        \toprule
          \multirow{2}{*}[0pt]{Settings}
  &\multicolumn{2}{c}{mIoU(\%)}  \\
  &nuScenes&SemanticKITTI\\
  \midrule
    manual dictionary &17.6 &18.9\\
    word2vec \cite{church2017word2vec}&30.4 &27.9 \\
    
        \bottomrule
       \end{tabular}
       \caption{ The label quality of maintaining a manual dictionary and calculating semantic similarity with word2vec in two datasets.}
        \label{tab:nusceneslabel}
\end{table}

In CLGM, we use the corresponding 2D image captures of two outdoor scene datasets as the source of our supervision signals. With our proposed 2D-PLGE, combined with FastSAM \cite{zhao2023fast} and SSA-Engine \cite{chen2023semantic}, we can quickly generate category-agnostic masks while obtaining semantic descriptions for each mask. For SSA-Engine, we have modified the source of each image patch instance mask required for inference, no longer using SAM but using FastSAM to significantly reduce our time. SSA-Engine generates the highest possible semantic description for each mask along with three alternative semantic descriptions. 
Subsequently, we use word2vec \cite{church2017word2vec} to calculate the semantic similarity between the four semantic descriptions in order of priority and the category text of the corresponding dataset, choosing the semantic category that appears most frequently among the semantic similarities calculated from the four semantic descriptions as the final pixel-level category. For comparison, we also manually maintained a dictionary following OpenScene \cite{peng2023openscene} as shown in Tab.\ref{tab:nusceneslabelmap} and Tab. \ref{tab:kittilablemap},  matching the most frequent words in the semantic descriptions with the category text of the dataset. Experiment results, as shown in Tab.~\ref{tab:nusceneslabel}, indicate that the label quality is higher when we use word2vec to calculate semantic similarity.

\begin{table}[ht]
\setlength{\tabcolsep}{0.4mm}

\centering

        \begin{tabular}{c|c|c}
        \toprule
         Ablation Target&Settings& mIoU (\%)   
     \\
        
        \midrule
        \multirow{3}{*}[0pt]{derivative threshold}&0.99&22.5\\
        &0.5&19.5\\
        &0.9&25.7\\
        \midrule
        \multirow{3}{*}[0pt]{winner portion}
        &2&23.4\\
        &3&25.7\\
        &4&19.0\\
        
        \bottomrule
       \end{tabular}
       \caption{The impact of the module hyperparameters.}
        \label{tab:appendix_hyperparameters}
\end{table}

In ANC, we let each sample record its training IoU for each round to fit its learning curve, noting that the training IoU is calculated from the prediction results and noisy labels. During the training process, we do not use any ground truth. The threshold for the derivative is set to 0.9. Following 
\cite{liu2022adaptive}, we also selected the optimal threshold to determine the appropriate correct time as Tab.~\ref{tab:appendix_hyperparameters} shows. We set the number of historical rounds to 5 for performance and memory considerations. In the frequency vector of the corrector, we set the best ratio $\omega$ for the winner label selection to 3. As Tab.~\ref{tab:appendix_hyperparameters} shows, too few ‘$\omega$’ settings may result in insufficient refinement, while too many ‘$\omega$’ settings can lead to chaotic category predictions. 

To obtain instance information under the sample, we use Patchwork++ \cite{lee2022patchworkpp} to remove the ground. For non-ground points, we set the block size to [10, 10, z] and slide to divide the point cloud with a stride of 10 to save time for the clustering algorithm. For DBSCAN \cite{schubert2017dbscan}, we set eps to 0.6 and min points to 5 to get complete point cloud instances.
\section{Visualization Results}
\begin{figure*}[hbp]
  \centering
    \includegraphics[width=0.9\linewidth]{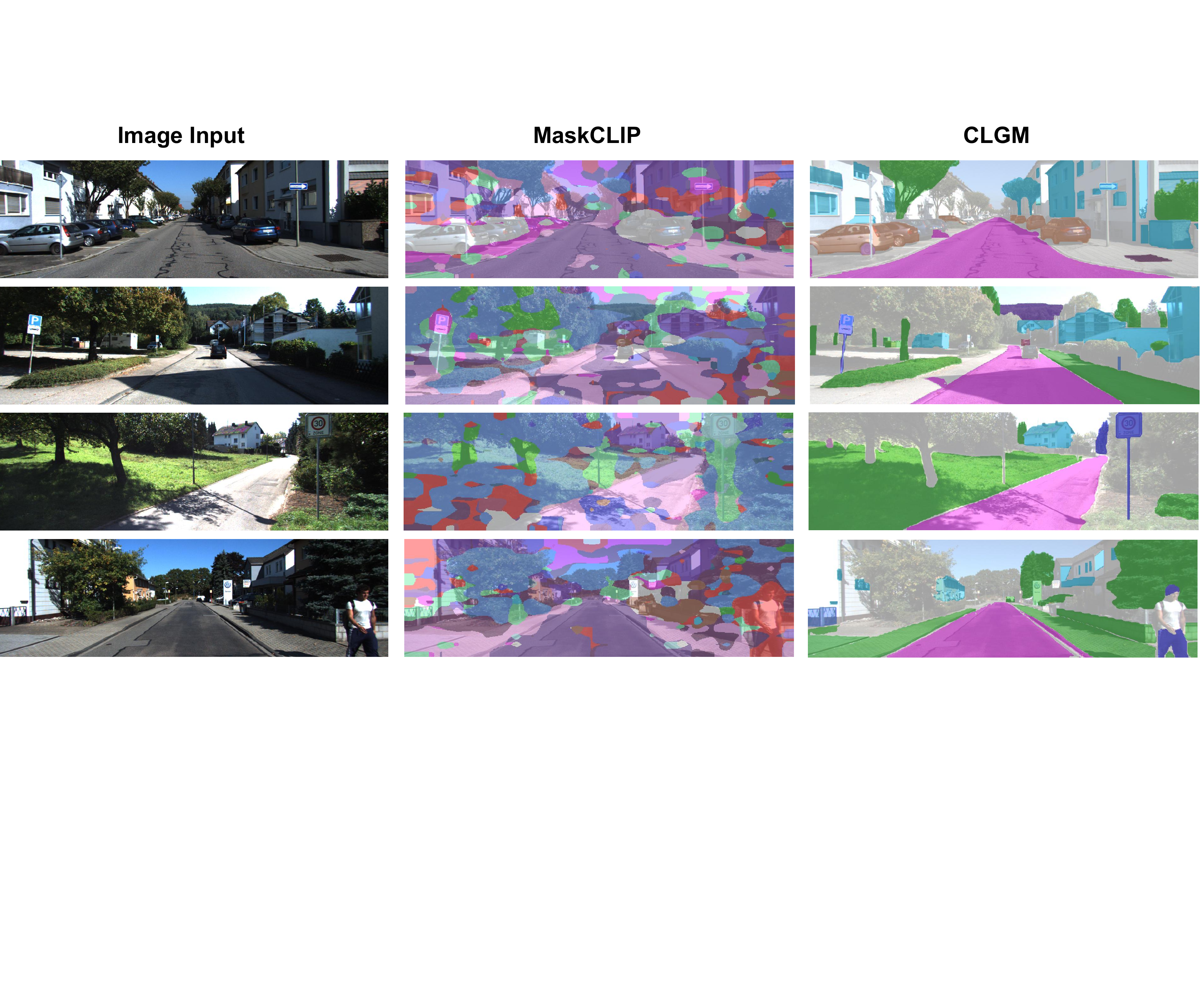}
  \caption{Qualitative results of 2D semantic labels generated in SemanticKITTI.  }
  \label{fig:kitti2dqualitative}
\end{figure*}

\begin{figure*}[htbp]
  \centering
 
    \includegraphics[width=0.9\linewidth]{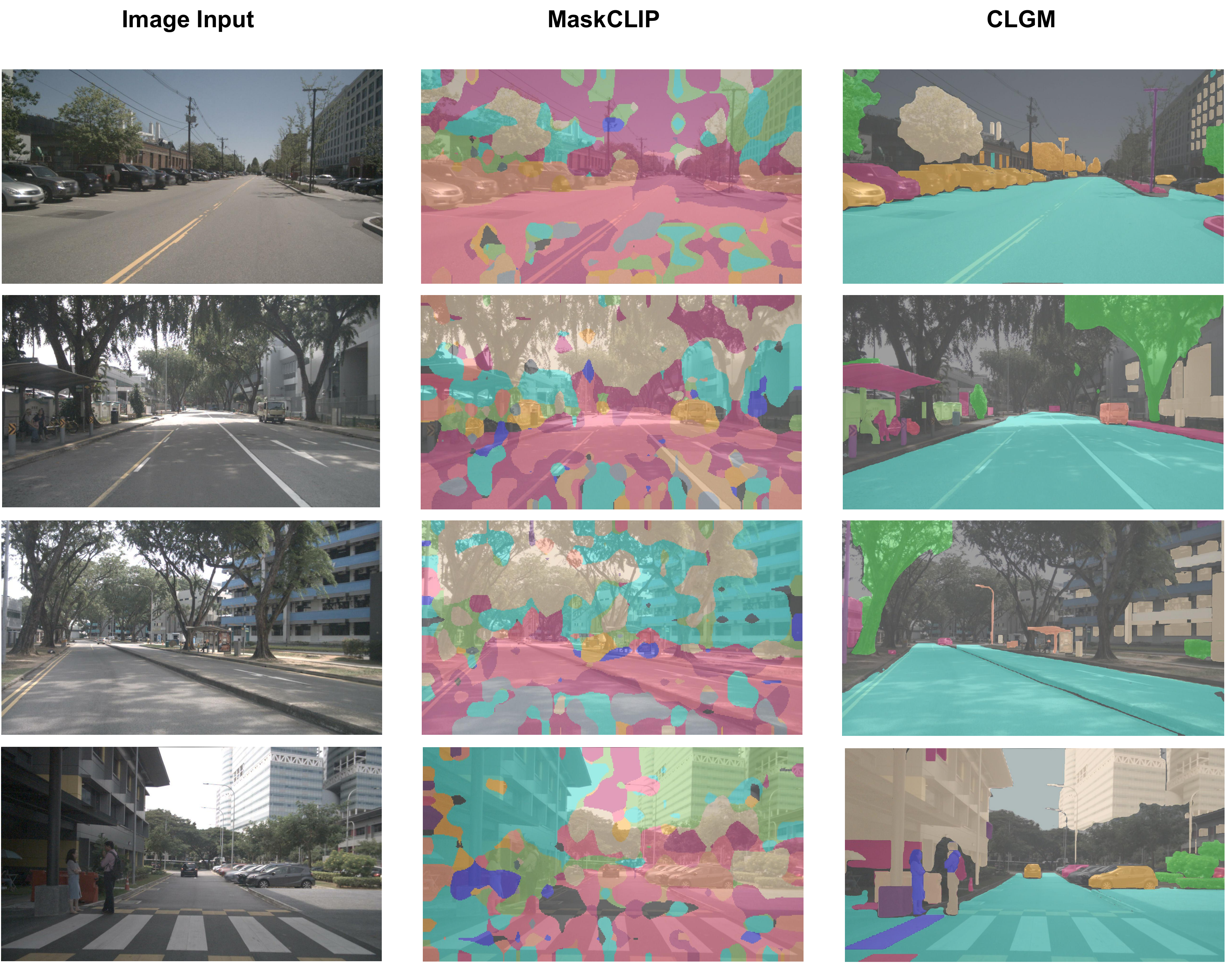}
  
  \caption{Qualitative results of 2D semantic labels generated in nuScenes. }
  \label{fig:nuscenes2dqualitative}
\end{figure*}

\begin{figure*}[htbp]
  \centering

    \includegraphics[width=\linewidth]{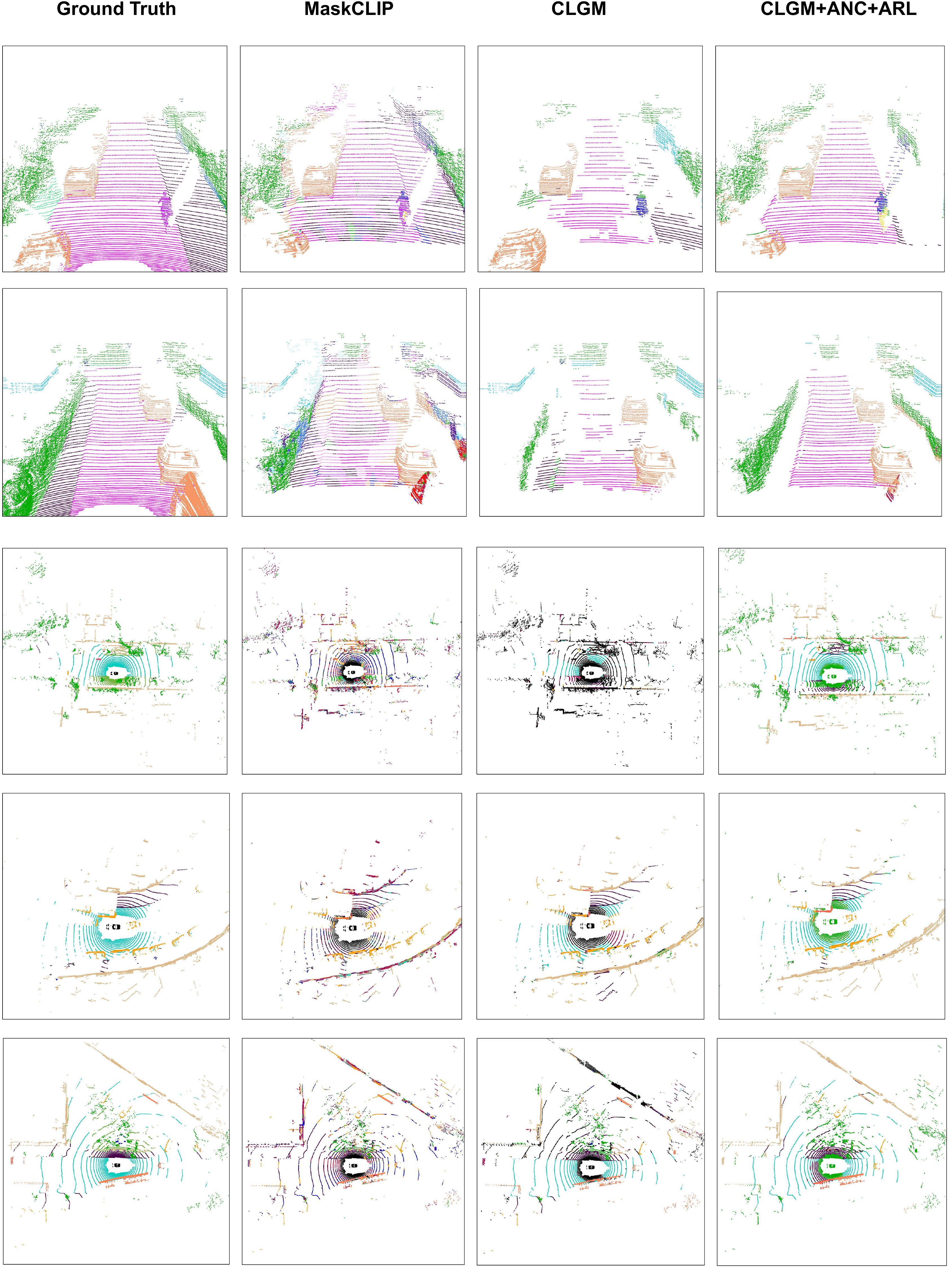}
 
  \caption{Qualitative results in SemanticKITTI (Row 1-2) and nuScenes (Row 3-5). }
  \label{fig:appqualitative}
\end{figure*}

We visualize our 2D image semantic label in SemanticKITTI and nuScenes datasets as shown in Fig.\ref{fig:kitti2dqualitative} and Fig. \ref{fig:nuscenes2dqualitative}. 
Our CLGM demonstrates better instance segmentation and semantic understanding capabilities on images. We also visualize more network prediction results in SemanticKITTI and nuScenes Datasets. Our ANC refurbished the noisy labels generated by CLGM, the ARL forces network fitting to cleaner samples as shown in Fig.\ref{fig:appqualitative}.
\newpage


\begin{table}[htb]

\centering

        \begin{tabular}{c|p{5cm}}
        \toprule
          nuScenes 16 labels
  &Our pre-defined labels  \\
        
        \midrule
        barrier & barrier, barricade\\
        bicycle & bicycle, bike \\
        bus & bus \\
        car & car \\
        construction vehicle & bulldozer, excavator, concrete mixer, crane, dump truck, vehicle, caravan, on rails\\
        motorcycle & motorcycle, motor\\
        pedestrian & pedestrian, people, person, child, man, woman\\
        traffic cone & traffic cone\\
        trailer & trailer, semi-trailer, cargo container, shipping container, freight container \\
        truck & truck\\
        driveable surface & road,highway\\
        other flat & curb, traffic island, traffic median\\
        sidewalk &sidewalk\\
        terrain & grass, grassland, lawn, meadow, turf, sod\\
        manmade & building, wall, pole, awning, door, city\\
        vegetation & tree, trunk, tree trunk, bush, shrub, plant, flower, woods\\
        \bottomrule
       \end{tabular}
       \caption{Label mapping for nuScenes 16 classes.}
        \label{tab:nusceneslabelmap}
\end{table}
\begin{table}[htb]
\centering
        \begin{tabular}{c|p{4cm}}
        \toprule
          SemanticKITTI 19 labels
  &Our pre-defined labels  \\
        
        \midrule
        car & car \\
        bicycle & bicycle, bike \\
        barrier & barrier, barricade\\
        motorcycle & motorcycle, motor\\
        truck & truck\\
        other-vehicle &bus, bulldozer, excavator, concrete mixer, crane, dump truck, vehicle, caravan, on rails, trailer, semi-trailer, cargo container, shipping container, freight container\\
        bicyclist & bicyclist\\
        motorcyclist & motorcyclist\\
        road & road,highway\\
        parking & parking\\
        sidewalk & sidewalk\\
        other-ground & curb, traffic island, traffic median, ground, street\\
        building & building, wall, manmade, awning, door, city\\
        fence & fence\\
        vegetation & tree, bush, shrub, plant, flower, woods\\
        trunk & trunk, tree trunk\\
        terrain & grass, grassland, lawn, meadow, turf, sod\\
        pole & pole\\
        traffic-sign & traffic cone, sign, traffic-sign\\
        \bottomrule
        
       \end{tabular}
       \caption{Label mapping for SemanticKITTI 19 classes.}
        \label{tab:kittilablemap}
\end{table}
\clearpage

\bibliography{aaai25}

\end{document}